\documentclass[letterpaper]{article} 
\usepackage{aaai23}  
\usepackage{times}  
\usepackage{helvet}  
\usepackage{courier}  
\usepackage[hyphens]{url}  
\usepackage{graphicx} 
\usepackage{natbib}  
\usepackage{caption} 
\usepackage{algorithm}
\usepackage{algorithmic}
\usepackage{newfloat}
\usepackage{listings}
\DeclareCaptionStyle{ruled}{labelfont=normalfont,labelsep=colon,strut=off} 
\floatstyle{ruled}
\newfloat{listing}{tb}{lst}{}
\floatname{listing}{Listing}
\usepackage[inline]{enumitem}
\usepackage{subfigure}
\usepackage{tabularx}
\usepackage{multirow}
\usepackage{relsize}
\usepackage{mathrsfs}
\usepackage{amsfonts}
\usepackage{amsmath}
\usepackage{booktabs} 
\DeclareMathAlphabet{\mathpzc}{OT1}{pzc}{m}{it}
\usepackage[toc,page]{appendix}

\title{Where Will Players Move Next? Dynamic Graphs and Hierarchical Fusion for Movement Forecasting in Badminton}

\author{
    Kai-Shiang Chang, Wei-Yao Wang, Wen-Chih Peng
}
\affiliations{
    National Yang Ming Chiao Tung University, Hsinchu, Taiwan\\
    \{kevin5260523.cs05, sf1638.cs05\}@nctu.edu.tw, wcpeng@cs.nycu.edu.tw
}

\begin{document}

\maketitle

\begin{abstract}
Sports analytics has captured increasing attention since analysis of the various data enables insights for training strategies, player evaluation, etc.
In this paper, we focus on predicting what types of returning strokes will be made, and where players will move to based on previous strokes.
As this problem has not been addressed to date, movement forecasting can be tackled through sequence-based and graph-based models by formulating as a sequence prediction task.
However, existing sequence-based models neglect the effects of interactions between players, and graph-based models still suffer from multifaceted perspectives on the next movement.
Moreover, there is no existing work on representing strategic relations among players' shot types and movements.
To address these challenges, we first introduce the procedure of the \textbf{P}layer \textbf{M}ovements (PM) graph to exploit the structural movements of players with strategic relations.
Based on the PM graph, we propose a novel \textbf{Dy}namic Graphs and Hierarchical Fusion for \textbf{M}ovement \textbf{F}orecasting model (DyMF) with interaction style extractors to capture the mutual interactions of players themselves and between both players within a rally, and dynamic players' tactics across time.
In addition, hierarchical fusion modules are designed to incorporate the style influence of both players and rally interactions.
Extensive experiments show that our model empirically outperforms both sequence- and graph-based methods and demonstrate the practical usage of movement forecasting\footnote{Code is available at https://github.com/wywyWang/CoachAI-Projects/tree/main/Movement\%20Forecasting.}.
\end{abstract}
\section{Introduction}
\label{sec:introduction}

In recent years, the rapid advancement of technology has brought great convenience for data collection and has increased data diversity, which has revolutionized sports analytics for conducting various applications.
For example, the National Basketball Association (NBA) has installed multiple cameras on every court to collect granular data on players’ movements.
These data are used to analyze the different views for recommending winning strategies, avoiding player injury, and scouting \cite{nba_example}.
On the other hand, the size of the sports analytics market is expected to expand 21.3\% from 2021 to 2028 \cite{sports-analytics-market}.
Evidently, effective analysis benefits the promising performance of teams or individuals on the field \cite{Wang2020badminton, melton2021examining, qian2022watching}, which has not only brought about the growing demand of sports analytics, but has also drawn researchers' attention to the flourishing and profitable sports market.

In this paper, we focus on turn-based sports and use badminton as the demonstration example.
Various studies related to badminton have been carried out to collect badminton data, such as retrieving information from videos \cite{DBLP:conf/apnoms/HsuWLCJIPWTHC19,DBLP:journals/eswa/CaoLSCL21,Yoshikawa2021} or sensors \cite{DBLP:journals/sensors/ChiuTST20,DBLP:journals/sensors/SteelsHFPPP20}.
Based on the collected data, several applications have been developed to analyze tactic performance \cite{DBLP:conf/cikm/Wang22}, e.g., shot influence \cite{DBLP:conf/icdm/WangCYWFP21,10.1145/3551391}, an analysis system \cite{DBLP:journals/tvcg/ChuXYLXYCZW22}, and stroke forecasting \cite{DBLP:conf/aaai/WangSCP22}.
However, in badminton, as a fast-paced sport, in addition to returning strokes, players' movements on the court are also a vital factor of simulating tactics, which has not been addressed in prior works.
Players with effective movements can reach the shuttle faster before it drops below the net, which enables them to return an aggressive stroke and take the dominant position \cite{badminton-footwork-and-position-on-the-court}.
Therefore, players and coaches can take advantage of movement forecasting since the movements on the court reflect the physical ability and the defensive position habit.
Players can therefore be trained to be familiar with where to catch the shuttle and where to defend based on the specific situation\footnote{Related work is in Appendix A due to space limitation.}.

\begin{figure}[!]
    \centering
    \includegraphics[height=!, width=0.40\textwidth,keepaspectratio=true]
    {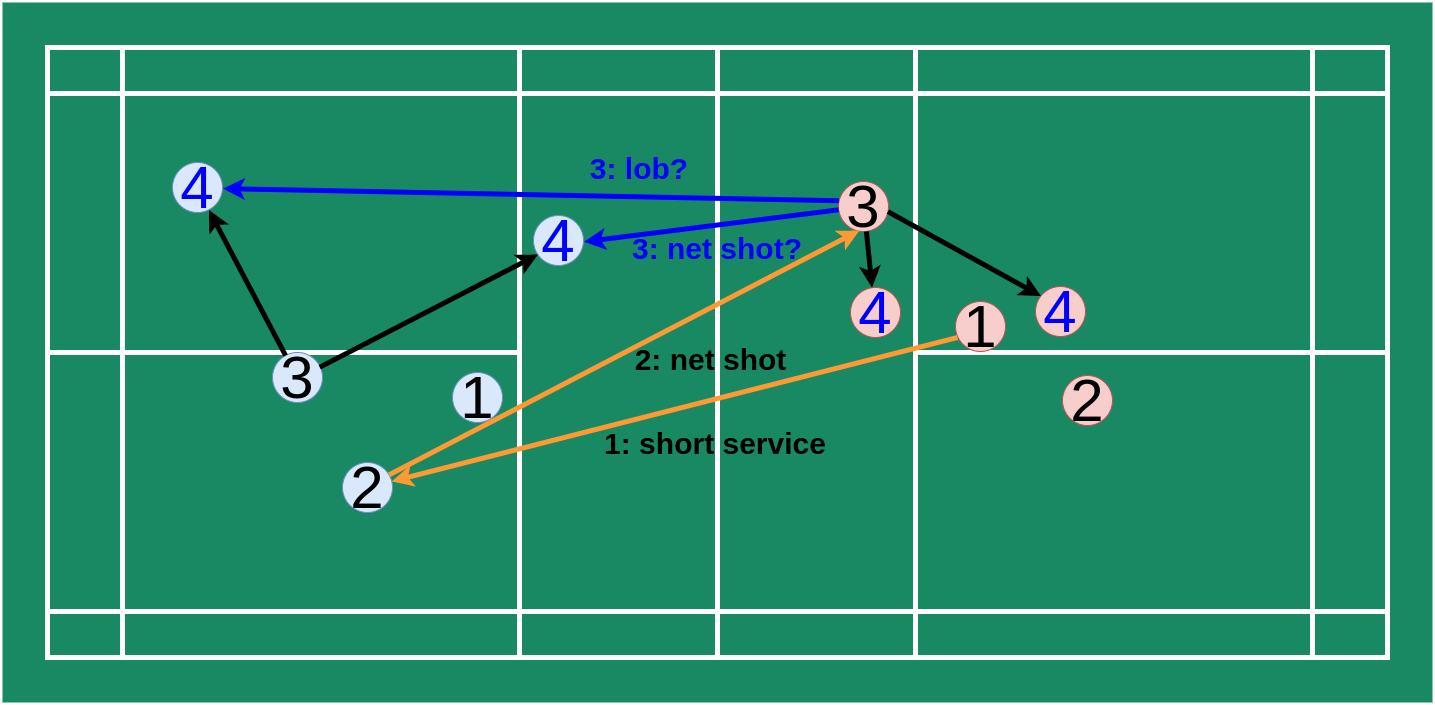}
    \caption{An example of movement forecasting. The blue nodes and red nodes represent the locations of two players at each time step, respectively. Nodes with blue numbers and blue lines are the possible choices of the moving locations of the two players and the shot type at the next time step.}
    \label{fig:schematic}
\end{figure}

Figure \ref{fig:schematic} illustrates an example of movement forecasting.
Given the locations of two players and the shot type they performed in a rally, the player on the right has several options of shot type and location to return the shot and move back to defense, while the player on the left also has multiple choices of location to catch the shuttle based on the location the opponent is returning to.
For example, when the left player returns a net shot to the right court, the right player can either also return a net shot and take a defense position closer to the net to prepare for another net shot, or can return a lob to the back court and move to the center of the half court to prepare for a smash.
In general, movement forecasting can be viewed as the more challenging task since the goal consists of not only stroke forecasting \cite{DBLP:conf/aaai/WangSCP22}, but also the movements of players.

To address this problem, we formulate movement forecasting as a sequence prediction task.
One potential solution is the sequence-to-sequence models, which have been widely applied for trajectory prediction \cite{DBLP:conf/icpr/GiuliariHCG20,DBLP:conf/aaai/ShiSFJZGWYS20,DBLP:conf/cvpr/ShafieePE21}.
However, in our task, sequence-to-sequence models are ineffective in terms of modeling the interactions between different players and the relations between the same player at different time steps, which has a great impact on the distance traveled and the destination.
For example, a net shot causes the opponent to move toward the net; consequently, the player has less time to go back to defense due to the short flying time.
On the other hand, a clear moves the opponent to the back court, which enables them to have sufficient time to move back to defense.
In addition, the meanings of moving between returning the shuttle and defending are different: when returning the shuttle, the player tends to move to the corner, and when defending, the tendency is to move to the center \cite{valldecabres2020players}.

To mitigate the aforementioned limitations, a badminton rally can be represented as a graph with various relations to describe different purposes of the movements across different time steps.
Nonetheless, there are three key challenges that hinder the direct use of existing graph methods in movement forecasting.
1) \textbf{Graph construction.}
Since previous approaches have not adopted graphs for turn-based sports analysis, there is no existing method for constructing badminton rallies into graphs.
Therefore, it is challenging to construct a graph to represent rally information, including the relations among players and their strokes. 
2) \textbf{Dynamic tactics.}
Players often change their tactics within a match or even a rally to avoid their strategies being seen through. 
In view of the ever-changing circumstances on the court, it is inadequate to fixedly model the styles and tactics of both players in different rallies and within the rally.
3) \textbf{Multifaceted aspects.}
To decide the next action, players consider not only their tactical preferences but also those of their opponents.
Furthermore, players are confronted with different significance between styles and rally progress according to the current circumstance.

In light of the above challenges, we propose a novel \textbf{Dy}namic graphs and hierarchical fusion for the \textbf{M}ovement \textbf{F}orecasting model (DyMF) to capture player information through different relations for predicting the next move of both players and the corresponding shot type.
For the first issue, we introduce a badminton graph called the \textbf{P}layer \textbf{M}ovements (PM) graph to construct strategic relations between two players and the players themselves in a rally across time.
Specifically, DyMF is an encoder-decoder architecture with interaction style extractors and hierarchical fusions.
For the second issue, the interaction style extractor is introduced to dynamically capture mutual interactions of players themselves and between players by aggregating information through different relations and pattern-generated weights.
For the third issue, we design a hierarchical fusion to take the style influence of both players into account and combine player-player interactions with rally interactions for predicting the next move.

To summarize, the main contributions of this paper are four-fold:
\begin{enumerate}
  \item We attempt to formalize the task of predicting the players' future movements and shot type based on the previous strokes. To the best of our knowledge, this is the first work for movement forecasting in badminton, which is also suitable for other turn-based sports such as tennis.
  \item To represent a rally as a graph, we introduce a graph construction method by transforming rallies into PM graphs with strategic relations between players' locations.
  \item Based on the PM graphs, we propose DyMF to extract different facets of interactions by interaction style extractors, and integrate them by hierarchical fusions.
  \item We conduct extensive experiments to demonstrate the effectiveness of our proposed model by comparing with state-of-the-art sequence- and graph-based methods on a real-world badminton dataset. Furthermore, an application scenario is illustrated for movement forecasting.
\end{enumerate}
\begin{figure*}
    \centering
    \includegraphics[height=!,width=0.98\linewidth,keepaspectratio=true]
    {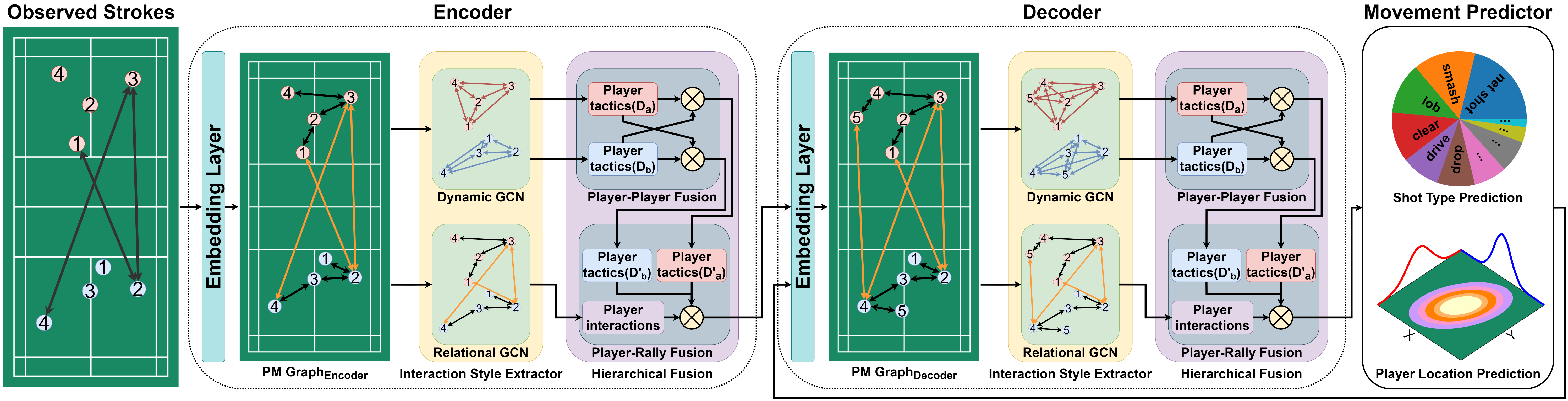}
    \caption{The architecture of our proposed DyMF. Interaction style extractors retrieve the context of mutual interactions on the PM graph and the dynamic tactics of a single player with previous patterns. Hierarchical fusions fuse the style influence of both players and rally interactions, and the movement predictor utilizes fused information to predict the locations of both players and the shot types they performed.
    }
    \label{fig:model-archtecture}
\end{figure*}

\section{Problem Formulation}
\label{sec:problems}

\sloppy
A badminton rally is denoted as $R=\{[S^{1}, \cdots, S^{|R|}], P\}$, where $S^{t}$ represents the $t$-th stroke, $P=({p_{a}}, {p_{b}})$ denotes the two players who participate in the rally and take turns hitting the shuttle to form a rally (i.e., $p_{a}$ is the player who serves and $p_{b}$ is the other player in a rally), and $|R|$ is the length of the rally.
At $t$-th time step (stroke), each stroke consists of two components $S^{t}=\{{L^{t}}, {s^{t}}\}$, where $L^{t} = \{{L^{t}_{a}}, {L^{t}_{b}}\}$ is the player locations of $p_{a}$ and $p_{b}$ when either player hits the shuttle, and $s^{t}$ represents the shot type played by either player.
Specifically, $L^{t}_{a} = (l^{t}_{ax}, l^{t}_{ay})\in \mathbb{R}^{2}$ and $L^{t}_{b} = (l^{t}_{bx}, l^{t}_{by})\in \mathbb{R}^{2}$ denote the 2D coordinates of the locations of two players on the court, respectively.
The movement forecasting task is defined as follows:
For each badminton rally, given the observed stroke sequence $[\{L^{1}, s^{0}\}, \cdots, \{L^{\tau}, s^{\tau-1}\}]$ with length $\tau$ and pairs of players $P$, the objective is to predict future strokes in the rally $[\{L^{\tau+1}, s^{\tau}\}, \cdots, \{L^{|R|}, s^{|R|-1}\}]$, including the locations of the two players and the shot types performed by either player.
We note that ${s}$ is offset by one index as the target is to predict the locations of the two players after each stroke; thus the first shot type $s^{0}$ is padded with a zero.

\section{Our Approach}
\label{sec:method}

Figure \ref{fig:model-archtecture} shows the overall architecture of our proposed DyMF.
To represent the badminton rally as a graph, the PM graph is designed with the locations of players at each time step and the relations between them across time.
Based on the PM graph, DyMF adopts an encoder-decoder architecture, where the encoder encodes the observed stroke sequence, while the decoder recurrently generates the locations of players and the shot types they performed based on the encoder contexts.
Specifically, both the encoder and decoder are composed of two modules, respectively, an interaction style extractor, and a hierarchical fusion.
The interaction style extractor takes players and corresponding locations as the information to capture mutual interactions of the players themselves and between both players by modeling the relations across time and the dynamic tactics of an individual player with past movement patterns.
To adapt to tactics and styles changing as the rally progresses, a hierarchical fusion is proposed to consider the style influence of both players and rally interactions based on the current circumstance. 

\begin{figure*}
    \centering
    \includegraphics[height=!, width=0.88\linewidth,keepaspectratio=true]
    {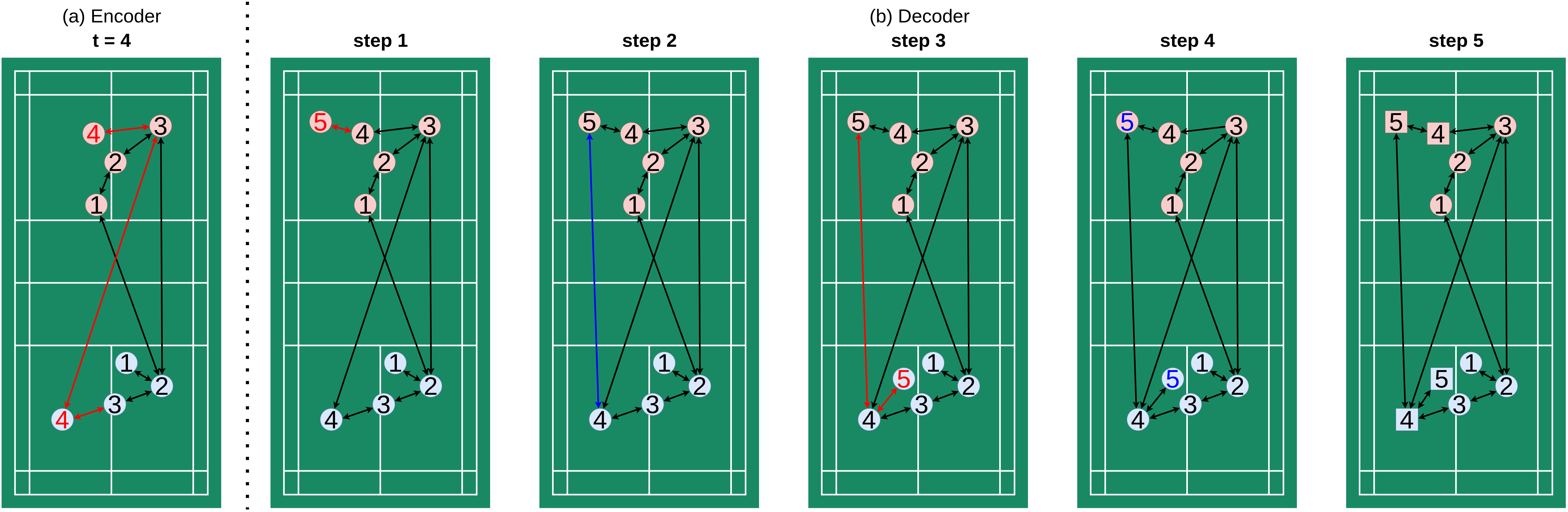}
    \caption{Illustrations of encoder and decoder PM graph constructions. The red nodes and blue nodes represent the locations of two players at the corresponding time steps, respectively. The newly added nodes and edges are decorated in red. The nodes with blue numbers and blue edges are the object to predict, and the square nodes represent the updated nodes at that time step.}
    \label{fig:graph-construction}
\end{figure*}

\subsection{Graph Construction}
\label{sec:graph-construction}

We first introduce the components in the PM graph and then illustrate in detail the graph construction procedure of the PM graph in the encoder and decoder.

\subsubsection{\textbf{PM Graph.}}
As a rally is comprised of players hitting the shuttle back and forth, the interactions between players within the rally can be described as structural representations in a graph.
However, directly applying existing graph construction methods (e.g., a complete graph) to rally sequences fails to model strategic relations between players.
To this end, a graph construction procedure is introduced by representing the rally progress as the evolution of a series of undirected multi-graphs $\mathpzc{G} = (\mathpzc{V}, \mathcal{E}, \mathpzc{R})$.
The nodes in the multi-graphs are denoted as $(v^{t}_{a}, v^{t}_{b}) \in\mathpzc{V}$, where $v^{t}_{a}$ and $v^{t}_{b}$ represent the locations of $p_a$ and $p_b$ at the $t$-th time step, respectively. 
The edges $(v^{i}_{a|b}, r, v^{j}_{a|b})\in\mathcal{E}$ represent the relation between the nodes of different players or the nodes of the same player between time steps $i$ to $j$, ${a|b}$ represents that the term is related to either ${p_a}$ or ${p_b}$, and the relations $r\in\mathpzc{R}$ consist of 12 strategic types listed as follows: \textbf{10 shot types: }net shot, lob, defensive shot, smash, drop, push/rush, short service, clear, drive, and long service. \textbf{Defend: }Link the nodes of the same player at the time steps $t-1$ and $t$ when the purpose of the movement is to return to the defense position. \textbf{Return: }Link the nodes of the same player at the time steps $t-1$ and $t$ when the purpose of the movement is to return the shuttle. 

\subsubsection{\textbf{Encoder Graph.}}
Figure \ref{fig:graph-construction} (a) illustrates an example of the graph construction procedure with the observed sequence with $\tau=4$ strokes.
At the first time step, there are two nodes that represent the initial locations of the two players $(p_{a}, p_{b})$ before the rally starts.
Afterwards, at each time step, two nodes and three edges are added to the graph to represent the movements and interactions of players on the court.
Specifically, the new nodes are the locations of the two players after either of them returns the shuttle.
The three new edges contain one shot type that links the node of the player who returns the shuttle with the time step $t-1$ and the node of the player who will return the shuttle at the next time step with the time step $t$ (red 3 to blue 4), one defend type that links the nodes of the player who returns the shuttle with the time steps $t-1$ and $t$ (red 3 to red 4), and one return type that links the nodes of the player who will return the shuttle at the next time step with the time steps $t-1$ and $t$ (blue 3 to blue 4).
The encoder graph is passed to the proposed encoder to learn the embeddings of the nodes.

\subsubsection{\textbf{Decoder Graph.}}
In the decoder stage, the PM graph is divided into five steps at each time step based on the encoder graph, which is depicted as an example at time step $t=5$ in Figure \ref{fig:graph-construction} (b):
1) Add one new node and one new edge, where the new node represents the location of the player who will return the shuttle at the next time step and the new edge is a return type.
2) Compute the node embeddings through the model decoder and predict the shot type of the stroke at the current time step by using the node embedding of the player who returns the shuttle with time step $t-1$ and the node embedding of the player who will return the shuttle at the next time step with time step $t$.
3) Then, a new node and two new edges are added in the decoder graph, where the new node represents the location of the player who returns the shuttle and moves back to the defense position, and the two new edges are a defense type (blue 4 to blue 5) and a shot type (blue 4 to red 5) based on the prediction from the previous step.
4) Compute the node embeddings through the model decoder and predict the locations of the two players based on the node embeddings of the last nodes of the two players.
5) Update the node embeddings of the last four nodes that participate in the graph construction at this time step by the model.
The decoder graph construction is then constructed recurrently until the end of the rally.


\subsection{Embedding Layer}
\label{sec:embedding-layer}
To represent the nodes with the locations of the players and corresponding players' information, the embedding layer takes $\{L^{t}_{a}, L^{t}_{b}\}$ and $({p_{a}}, {p_{b}})$ as input to calculate the initial representation of the node $e^{t}_a$ and $e^{t}_b$ of $v^{t}_a$ and $v^{t}_b$ at the $t$-th time step, respectively:
\begin{equation}
    \small
    e^{t}_{a|b} = W^e(ReLU(W^L L^{t}_{a|b}) \mathbin\Vert W^p p_{a|b}),
    \label{eq:embedding-layer}
\end{equation}
where $W^L \in \mathbb{R}^{d_l \times 2}$, $W^p \in \mathbb{R}^{d_p \times N_p}$, $W^e \in \mathbb{R}^{d_e \times (d_l + d_p)}$, are learnable matrices, and $\mathbin\Vert$ is the concatenation operator.
$d_l$, $d_p$ and $d_e$ denote the dimension of location, player, and node representations, respectively, and $N_P$ is the total number of players in the dataset.

It is noted that the initial representation in the decoder is employed with the locations of the two players at the last time step, since their locations will be predicted at the end of the time step.
We also share the parameters $W^L$ and $W^p$ in the embedding layers on the encoder and decoder to reduce the number of parameters similar to \cite{DBLP:conf/aaai/WangSCP22}.

\subsection{Interaction Style Extractor}
The interactions and style confrontations between the players on the court are necessary for making a decision on the next move for both players.
For example, a defensive player tends to return a clear or a lob, which gives more time to move to a better position for defense and wait for the unforced error by the opponent.
Therefore, the interaction style extractor is designed with relational GCN and dynamic GCN to capture the interactions between players and the individual style of both players, respectively.

\subsubsection{\textbf{Relational GCN.}}

Both the shot types by either player and the mutual interactions (i.e., purposes of the movements) play a vital role in organizing tactics to influence the next locations of both players.
For instance, a lob moves the player to the back court, while a net shot moves the player to the net.
Moreover, the player tends to be in the center of the court when the purpose of the movement is to defend.
When the purpose of the movement is to return the shuttle, the tendency is to move to the corner since the opponent would try to mobilize the player.
Therefore, relational GCN is introduced to capture the interactions between different players and the relations between the same player at different time steps.


Inspired by R-GCN \cite{DBLP:conf/esws/SchlichtkrullKB18}, the architecture of relational GCN is a propagation model that learns the node embedding by aggregating the information from neighbors based on the relations between them.
To preserve the information to the next layer, the message-passing process is added with a self-connection as:
\begin{equation}
    \small
    z^{i(l+1)} = \sigma(\mathlarger{\mathlarger{\Sigma}}_{r\in\mathpzc{R}}\mathlarger{\mathlarger{\Sigma}}_{j\in N^i_r}a^{i}_rW^{r(l)}z^{j(l)} + W^{o(l)}z^{i(l)}),
    \label{eq:relational-gcn}
\end{equation}
where $z^{i(l)}$ denotes the hidden state of the $i$-th node in $\hat v$ in layer $l$, where $\hat v = [v^{1}_a, v^{1}_b, \cdots, v^{t}_a, v^{t}_b]$, $N^i_r$ is a set of neighbors of $i$-th node in $\hat v$ for a corresponding relation $r$, $\sigma$ is the activation function, $W^{r(l)}, W^{o(l)} \in \mathbb{R}^{d_e \times d_e}$ are the learnable matrices in layer $l$, and $a^{i}_r$ is the normalized constant.
The hidden states of the nodes at the first layer in the encoder are the initial node embeddings $e^{i}_a$ and $e^{i}_b$.
Besides, the propagated node embeddings $\hat e^{t}_{a}$ and $\hat e^{t}_{b}$ at previous time steps are used for the hidden states at the first layer in the decoder, except that the two newly added nodes use the initial node representation.

To stabilize the training of each relation, we adopt the regularization method of basis decomposition \cite{DBLP:conf/esws/SchlichtkrullKB18} for $W^{r(l)}$ :
\begin{equation}
    \small
    W^{r(l)} = \mathlarger{\mathlarger{\Sigma}}^{B}_{b=1}c^{rb(l)}M^{b(l)},
    \label{eq:basis-decomposition}
\end{equation}
where $B$ is the number of basis matrices and $c^{rb(l)}$ denotes the learnable parameters of relation $r$ for the basis matrix $M^{b(l)} \in \mathbb{R}^{d_e \times d_e} $ in layer $l$.

\subsubsection{\textbf{Dynamic GCN.}}
\label{sec:dynamic-gcn}
\citet{DBLP:conf/aaai/WangSCP22} demonstrated the significance of considering player styles for predicting shot types and destination locations, but ignored the dynamic personal style of each player, which is frequently changed at each stroke for adjusting their tactics.
To address this issue, we propose dynamic GCN by generating the weights based on the players' patterns to reflect dynamic personal style.

\sloppy
Specifically, the player information is first included in the outputs of the embedding layer:
\begin{equation}
    \small
    n^{t}_{a|b} = W^n(e^{t}_{a|b} \mathbin\Vert p_{a|b}),
    \label{eq:dynamic-gcn-input}
\end{equation}
where $W^n \in \mathbb{R}^{d_e \times (d_e + d_p)}$ are learnable matrices.
In the decoder, $e^{t}_{a|b}$ is replaced by the propagated node representation $\hat e^{t}_{a|b}$.
Then, we apply a 1D convolutional neural network (Conv1D) with kernel size $K$ to the $E^{t}_{a|b}=[n^{1}_{a|b}, \cdots, n^{t}_{a|b}]$ to extract the local movement patterns of each player:
\begin{equation}
\small
\begin{aligned}
    C^{t}_{a|b} = Conv1D(E^{t}_{a|b}).
    \label{eq:1d-convolution}
\end{aligned}
\end{equation}
We note that $E^{t}_{a|b}$ is padded with zeros until the dimension is equal to $d_e$ and the same length of input as the corresponding output.

Afterwards, LSTM \cite{DBLP:journals/corr/SakSB14} is applied to the pattern sequences $C^{t}_{a|b}$ to capture the long-term patterns:
\begin{equation}
    \small
    h^{(s+1)}_{a|b} = LSTM(h^{(s)}_{a|b}, c^{(s)}_{a|b}, C^{t(s)}_{a|b}; W^{LSTM}),
    \label{eq:lstm}
\end{equation}
where $h^{(s)}_{a|b},c^{(s)}_{a|b} \in \mathbb{R}^{d_e}$ are the $s$-th hidden state and cell state.
$C^{t(s)}_{a|b} \in \mathbb{R}^{d_e}$ are the $s$-th pattern of $C^t_{a}$ and $C^t_{b}$, and $W^{LSTM}$ is the learnable parameters.
The initialization of the hidden state and cell state is filled with zeros following \cite{DBLP:conf/icdm/WangCYWFP21}.

\sloppy
Finally, the pattern-generated weights $Q_{a|b} = [h^{(1)}_{a|b}, \cdots, h^{(t)}_{a|b}]$ are regarded as the weights of dynamic GCN to calculate the context of player interactions for each node by replacing the weights in the graph convolutional network (GCN) \cite{DBLP:conf/iclr/KipfW17} with $Q_{a|b}$:


\begin{equation}
\small
\begin{aligned}
    d^{i}_{a|b} = GCN(n^{i}_{a|b}; Q_{a|b}),
    \label{eq:dynamic-gcn}
\end{aligned}
\end{equation}
where $d^{i}_{a|b}$ denotes the hidden states of nodes $v^i_a$ and $v^i_b$.

\subsection{Hierarchical Fusion}
\label{sec:hierarchical-fusion}
Although players move to the next location for returning the next stroke based on rally interactions and players' styles, it is expected that the importance of each varies under different circumstances.
That is, promising players change their tactics according to their opponents in order to seize the winning opportunity.
Hence, we propose a hierarchical fusion to decide the mutual importance of each style by modeling the style influence between the opponent and the current returning player, and to merge the contexts of rally interactions and player styles.

\sloppy 
\subsubsection{\textbf{Player-Player Fusion.}}
\label{sec:player-player-fusion}
Parallel co-attention \cite{DBLP:conf/nips/LuYBP16} is adopted in the sequence of the player's dynamic tactics from the output of dynamic GCN $D^{t}_a = [d^{1}_a, \cdots, d^{t}_a]$ and $D^{t}_b = [d^{1}_b, \cdots, d^{t}_b]$ to consider the mutual influence of the style $\hat a$ and $\hat b$ of each player:
\begin{equation}
\small
\begin{gathered}
    G = tanh((D^{t}_a)^TW^DD^{t}_b),\\
    H_a = tanh(W^aD^{t}_a + (W^bD^{t}_b)G^T), \\
    H_b = tanh(W^bD^{t}_b + (W^aD^{t}_a)G),\\
    att_{a|b} = softmax((w^{ha|hb})^TH_{a|b}), \\
    {\hat a|\hat b} = \mathlarger{\mathlarger{\Sigma}}^{t}_{i=1}att^i_{a|b} d^{i}_{a|b}, 
\end{gathered}
\end{equation}
where $W^D, W^a, W^b \in \mathbb{R}^{d_e \times d_e}$ are learnable matrices and $w^{ha}, w^{hb} \in \mathbb{R}^{d_e}$ are learnable parameters.
The outputs $\hat a$ and $\hat b$ are then transformed to the overall influence $f_a$ and $f_b$:
\begin{equation}
\small
\begin{aligned}
    f_a = sigmoid(w^{\hat a}\hat a), f_b = sigmoid(w^{\hat b}\hat b),
\end{aligned}
\end{equation}
where $w^{\hat a}, w^{\hat b} \in \mathbb{R}^{d_e}$ are learnable parameters.

Afterwards, we derive the context of each player's dynamic player's style with the opponent's influence:
\begin{equation}
    \small
    D^{'t}_{a|b} = f_{b|a} \times D^{t}_{b|a} + D^{t}_{a|b}.
\end{equation}

\subsubsection{\textbf{Player-Rally Fusion.}}
The player-rally fusion aims at capturing the importance of player-player styles and rally interactions to predict each player's next move.
To fuse these contexts, we first use the $t$-th player-player styles $d^{'t}_{a|b}$ from $D^{'t}_{a|b}$ and rally interactions in the last layer in the relational GCN $z^{t(l_{last})}_{a|b}$ of both players to determine the corresponding dynamic weights $\alpha_a, \alpha_b, \beta_a, \beta_b$:
\begin{equation}
\small
\begin{gathered}
    \alpha_{a|b} = sigmoid(w^Sd^{'t}_{a|b}),
    \beta_{a|b} = sigmoid(w^Zz^{t(l_{last})}_{a|b}),
\end{gathered}
\end{equation}
where $w^S, w^Z \in \mathbb{R}^{d_e}$ are learnable parameters. 

Afterwards, the propagated node embeddings $\hat e^t_a$ and $\hat e^t_b$ are updated by:
\begin{equation}
\small
\begin{aligned}
    \hat e^t_{a|b} = \alpha_{a|b} \times d^{'t}_{a|b} + \beta_{a|b} \times z^{t(l_{last})}_{a|b}.
\end{aligned}
\end{equation}

\subsection{Movement Predictor}
\label{sec:prediction}

To predict the shot type at the $t$-th time step (step 2 in the PM graph), the propagated node embeddings of both players at the $t$-th time step are calculated as:
\begin{equation}
    \small
    \hat s^t = W^s(\hat e^{t-1}_a\mathbin\Vert \hat e^{t}_b),
\end{equation}
where $W^s \in \mathbb{R}^{N_t \times 2d_e}$ are learnable matrices, and $N_t$ is the number of shot types.
We note that the former term becomes $\hat e^{t-1}_b$ and the latter becomes $\hat e^{t}_a$ when the number of the time step is odd, which switches the returning player and the defending player.

To predict the locations of the two players at the $t$-th time step (step 4 in the PM graph), we follow \cite{DBLP:conf/aaai/WangSCP22} to learn a bivariate Gaussian distribution for the location of each player.
Specifically, a bivariate Gaussian distribution consists of five parameters $\{\mu_x, \mu_y, \sigma_x, \sigma_y, \rho\}$, which represent the mean of the 2D coordinates, the variance of the 2D coordinates, and the corresponding correlation coefficient.
We apply a linear layer to the concatenation of the propagated node embeddings of two players to predict two bivariate Gaussian distributions $\hat {BG}^{t}_a$ and $\hat {BG}^{t}_b$ for both players:
\begin{equation}
    \small
    \{\hat {BG}^{t}_a, \hat {BG}^{t}_b\} = W^{BG}(\hat e^{t}_a \mathbin\Vert \hat e^{t}_b),
\end{equation}
where $W^{BG} \in \mathbb{R}^{10 \times 2d_e}$.
The dimension 10 is the total number of two five-parametric distributions.
The predicted coordinates of the $t+1$ time step for each player $L^{t+1}_{a}$ and $L^{t+1}_{b}$ are then sampled from $\mathcal{N} (\hat {BG}^{t}_a)$ and $\mathcal{N} (\hat {BG}^{t}_b)$, respectively.

The objective function is to minimize the total loss $\mathcal{L}$:
\begin{equation}
    \small
    \mathcal{L}=\mathcal{L}_{shot\_type} + (0.5 \times \mathcal{L}_{location\_a} + 0.5 \times \mathcal{L}_{location\_b}),
    \label{total_loss}
\end{equation}
where $\mathcal{L}_{shot\_type}$ is the cross-entropy loss:
\begin{equation}
    \small
    \mathcal{L}_{type} = - \mathlarger{\mathlarger{\Sigma}}_{R \in \mathcal{D}} \mathlarger{\mathlarger{\Sigma}}^{|R|-1}_{t=\tau} s^{t} log(softmax(\hat s^{t})),
    \label{cross-entropy-loss}
\end{equation}
and $\mathcal{L}_{location\_a}$ and $\mathcal{L}_{location\_b}$ are the negative log-likelihood losses for each player:
\begin{equation}
\small
\begin{gathered}
    \mathcal{L}_{location\_a} = - \mathlarger{\mathlarger{\Sigma}}_{R \in \mathcal{D}} \mathlarger{\mathlarger{\Sigma}}^{|R|}_{t=\tau+1} log(\mathcal{P}(l^{t}_{ax}, l^{t}_{ay} | \hat {BG}^{t}_a)),\\
    \mathcal{L}_{location\_b} = - \mathlarger{\mathlarger{\Sigma}}_{R \in \mathcal{D}} \mathlarger{\mathlarger{\Sigma}}^{|R|}_{t=\tau+1} log(\mathcal{P}(l^{t}_{bx}, l^{t}_{by} | \hat {BG}^{t}_b)),
    \label{negative-log-lokelihood-loss}
\end{gathered}
\end{equation}
where $\mathcal{D}$ is the set of rallies.
\section{Experiments}
\label{sec:experiments}

\subsection{Experimental Settings}
\subsubsection{\textbf{Badminton Dataset.}}
\label{badminton-dataset}
Since there is only one dataset of turn-based stroke records, we employ the badminton singles dataset \cite{DBLP:conf/aaai/WangSCP22}.
After filtering the rallies with any missing value, the badminton dataset has 75 matches played by 31 high-ranking players remaining, which contain 180 sets, 4,325 rallies, and 43,191 strokes.
Following \cite{DBLP:conf/aaai/WangSCP22}, we use the same 10 shot types, and the maximum rally lengths are truncated to 35.

For the task of movement forecasting, we utilize the rally id to separate rallies.
Each rally contains the ball round of each stroke, which consists of the players, their locations, and the shot type either of them performed.
The first 80\% of rallies are split as training data and the last 20\% as testing data for each match to have the model acquire each player's previous information.

\subsubsection{\textbf{Baseline Methods.}}
\label{sec:baseline-methods}
As this task has yet to be addressed, there are no existing baselines for direct comparison.
Therefore, we examine our model with several state-of-the-art sequence prediction methods to verify its effectiveness.
Specifically, these baselines are categorized into two groups: \textbf{Sequence-based models}: 1) Seq2Seq \cite{DBLP:conf/nips/SutskeverVL14}, 2) TF \cite{DBLP:conf/icpr/GiuliariHCG20}, and 3) ShuttleNet \cite{DBLP:conf/aaai/WangSCP22}. \textbf{Graph-based models}: 1) dNRI \cite{DBLP:conf/cvpr/GraberS20a}, 2) GCN\textsubscript{PM} \cite{DBLP:conf/iclr/KipfW17}, 3) R-GCN\textsubscript{PM} \cite{DBLP:conf/esws/SchlichtkrullKB18}, and 4) E-GCN\textsubscript{PM} \cite{DBLP:conf/aaai/ParejaDCMSKKSL20}.
It is worth noting that we introduce three new variants of the graph-based approaches (GCN\textsubscript{PM}, R-GCN\textsubscript{PM}, and E-GCN\textsubscript{PM}) with encoder-decoder structures and the proposed PM graph for better exploring the performance of our proposed method.
More details and setups of baselines are given in Appendix B.
 
\subsubsection{\textbf{Parameter Settings.}}
\label{sec:experiment-settings}
The representation dimension ($d_l, d_p, d_e$) is set to 16 for all models.
The dropout rate for each module is 0.1, and the number of heads in TF and ShuttleNet is set to 2.
E-GCN\textsubscript{PM}, R-GCN\textsubscript{PM}, GCN\textsubscript{PM} and our model have two GNN layers.
The number of layers of TF is 1.
The activation function $\sigma$ in R-GCN\textsubscript{PM}, GCN\textsubscript{PM}, E-GCN\textsubscript{PM} and our model applies ReLU at the first layer and Sigmoid at the second layer.
The normalized constants ($a^{i}_r$) for R-GCN\textsubscript{PM} and our model are set to 1 similar to \cite{DBLP:conf/esws/SchlichtkrullKB18} and the number of basis matrices ($B$) is 3.
The kernel size $K$ is set to 3 and the layer of LSTM is set to 1 in dynamic GCN.
We train all models for 100 epochs with the Adam optimizer, where batch size is set to 32 and learning rate is set to 0.001.
All the training and evaluation phases were conducted on a machine with Ubuntu 20.04, Intel i7-9700K 3.6GHz CPU, and Nvidia GTX 2070 8GB GPU, while the methods are implemented in Python 3.7 with the PyTorch 1.9.0.
5-fold cross-validation was conducted for tuning the hyper-parameters.

\subsubsection{\textbf{Evaluation Metrics.}}
Following \cite{DBLP:conf/aaai/WangSCP22} for evaluating stochastic models, we generate 10 sequences for each rally and take the closest one as the final prediction for evaluation, and utilize mean square error (MSE) and mean absolute error (MAE) for the prediction of the locations of the two players, and cross-entropy (CE) for the prediction of shot type.
All the results are the average of 5 different random seeds.

\subsection{Performance Comparison}
\label{sec:performance-comparison}

\begin{table*}[h]
    \small
    \centering
    \newcommand{\specialcell}[2][c]{%
  \begin{tabular}[#1]{@{}c@{}}#2\end{tabular}}

\begin{tabular}{m{25pt}c|ccc|ccc|ccc|c}
    \toprule
    & & \multicolumn{3}{c|}{$\tau=2$} & \multicolumn{3}{c|}{$\tau=4$} & \multicolumn{3}{c|}{$\tau=8$} & \multicolumn{1}{c}{}\\
    \cmidrule{3-12}
    Group& Model & \specialcell{MSE\\location} & \specialcell{MAE\\location} & \specialcell{CE\\shot type} & \specialcell{MSE\\location} & \specialcell{MAE\\location} & \specialcell{CE\\shot type} & \specialcell{MSE\\location} & \specialcell{MAE\\location} & \specialcell{CE\\shot type} & \specialcell{Avg.\\Rank} \\
    \midrule
    Sequence    & Seq2Seq                   & 1.3093 & 1.7552 & 2.0687 &
                                              1.1880 & 1.6739 & 1.9905 &
                                              1.2033 & 1.6920 & 1.9751 &
                                              4.8 \\
                & TF                        & 1.2233 & 1.7122 & 2.0915 &
                                              1.2206 & 1.7094 & 2.0258 &
                                              1.1569 & 1.6785 & 2.0134 &
                                              5.0 \\
                & ShuttleNet                & \underline{1.1345} & \underline{1.6179} & \underline{1.9599} &
                                              \underline{1.1539} & 1.6364 & \underline{1.9575} &
                                              \underline{1.1399} & \underline{1.6360} & \underline{1.9654} &
                                              \underline{2.1} \\
    \midrule
    Graph       & dNRI                      & 1.2803 & 1.7865 & 2.4269 &
                                              1.2994 & 1.7921 & 2.4214 &
                                              1.4651 & 1.9117 & 2.4318 &
                                              7.2 \\
                & GCN\textsubscript{PM}     & 1.2018 & 1.6878 & 1.9630 &
                                              1.2033 & 1.6964 & 1.9854 & 
                                              1.2676 & 1.7508 & 1.9875 &
                                              4.4 \\
                & R-GCN\textsubscript{PM}   & 1.1724 & 1.6329 & 2.0620 &
                                              1.1681 & \underline{1.6339} & 2.0587 &
                                              1.2006 & 1.6554 & 2.0470 &
                                              3.8 \\
                & E-GCN\textsubscript{PM}   & 2.8332 & 2.5003 & 2.2878 &
                                              11.0312& 3.5865 & 2.2786 &
                                              1314.0263 & 13.9290 &	2.1180 &
                                              7.7 \\
    \midrule
                & DyMF (Ours)              & \textbf{1.1006} & \textbf{1.5795} & \textbf{1.9543} &
                                             \textbf{1.1061} & \textbf{1.5888} & \textbf{1.9487} &
                                             \textbf{1.0827} & \textbf{1.5739} & \textbf{1.9570} &
                                             \textbf{1.0} \\
    \bottomrule
\end{tabular}
    \caption{Performance of baseline models and our proposed model. In terms of MSE, MAE, and CE in different observed stroke sequence lengths ($\tau$), the best results are highlighted in boldface and the second best are underlined.}
    \label{tab:performance}
\end{table*}

To comprehensively evaluate the performance of each method, we perform experiments on three different observed stroke sequences $\tau = 2$, $4$, and $8$.
We note that extensive ablation experiments are discussed in Appendix C.1.
Table \ref{tab:performance} reports the results of our proposed model and the baselines, which demonstrates that DyMF consistently surpasses all baselines in terms of all metrics and different observed lengths.
Quantitatively, the improvement in our model is up to 35.3\% for MSE, 21.5\% for MAE, and 24.3\% for CE.
We summarize the observations as follows:

1) Seq2Seq and TF degenerate location performance (MSE and MAE) compared to GCN\textsubscript{PM} and R-GCN\textsubscript{PM}, which not only indicates that naive sequence-to-sequence methods fail to model structural information between locations, but also demonstrates the importance of applying the PM graph to capture the influence on the intent of movement.
2) dNRI hinders all performance more substantially, although dynamically modeling the relation between players, which implies that two same nodes for representing two players separately across time is insufficient for the movement forecasting task.
This again raises the need for using a PM graph with new location nodes to predict the next move of each player.
3) As ShuttleNet is designed for turn-based sequences and fusing different aspects according to different strokes, we can observe that it performs better predictions than both the sequence-based and graph-based baselines.
Nonetheless, the comparison of DyMF and ShuttleNet reveals the importance of considering the purpose of players' movements.
Our proposed DyMF outperforms ShuttleNet for all scenarios, which is attributed to our interaction style extractors for modeling mutual interactions across time as well as the dynamic tactics based on the previous movement patterns, and the player-player fusion for integrating the styles of each player.
4) We note that E-GCN\textsubscript{PM} produces unfavorable performance, possibly due to the unstable states when there are only a few nodes in badminton rallies, and so is not suitable for this task.

\subsection{Case Study}
\label{sec:case-study}

\begin{figure}
    \centering
    \includegraphics[height=!, width=0.83\linewidth,keepaspectratio=true]
    {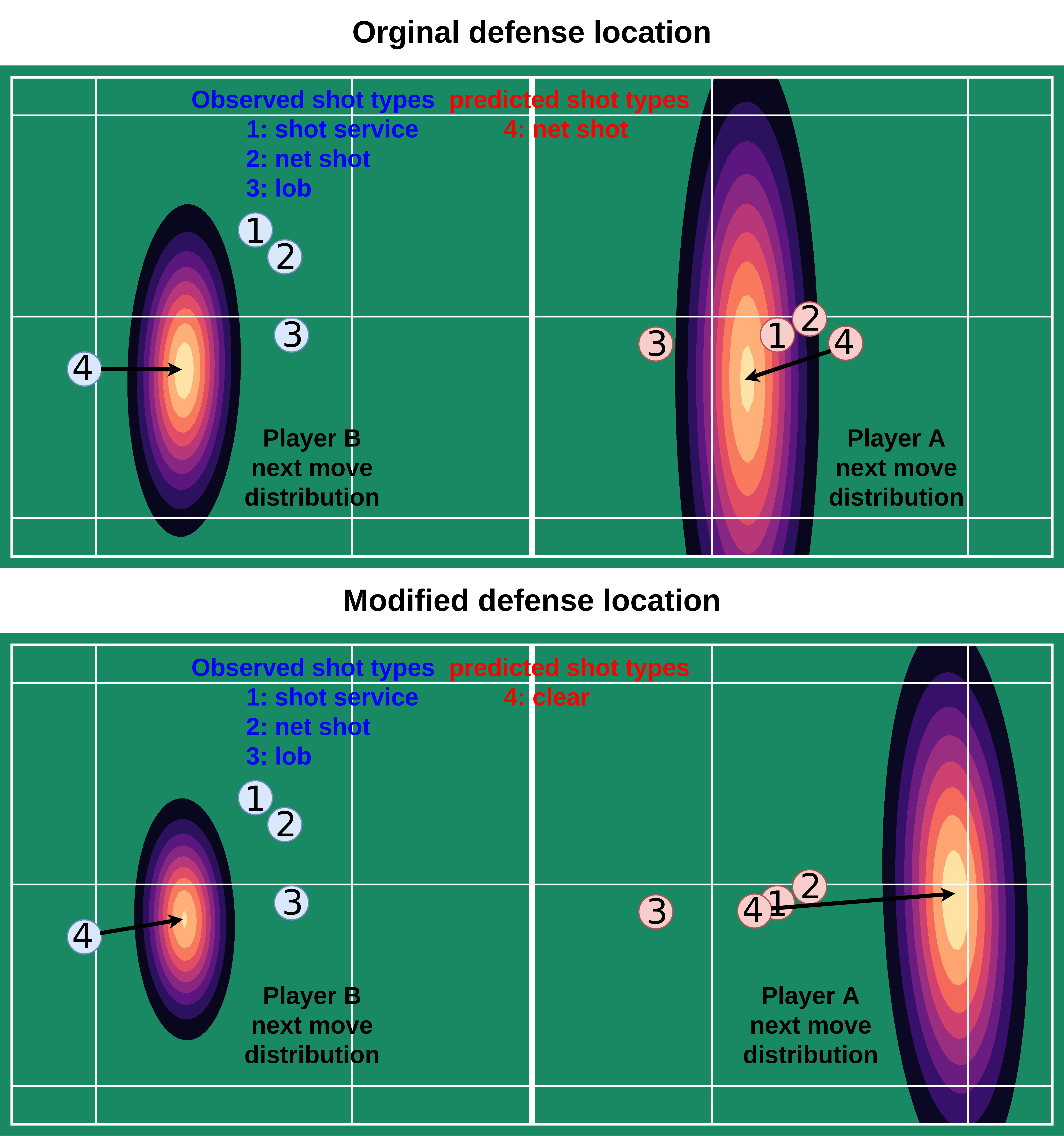}
    \caption{Illustrations of strategies in different defense locations. Red nodes are the served player (Player A) and blue nodes are the receiving player (Player B).}
    \label{fig:case-study-different-defense}
\end{figure}

The analysis of the players' movements on the court can be viewed as understanding behaviors of players, where stroke forecasting \cite{DBLP:conf/aaai/WangSCP22} only focuses on what types of strokes, and where to return to.
One potential technique to investigate this is to use statistical methods on historical records.
However, the occurrence of area coordinates is too sparse to be analyzed.
To that end, a paradigm is illustrated with DyMF to investigate the practicality of movement forecasting, and more cases are discussed in Appendix C.2.
Figure \ref{fig:case-study-different-defense} shows a scenario where Player A takes a defense location toward the net after returning the lob (red 3) and moving back to the middle court (red 4).
In this case, we aim to change the defense location of the player to analyze the reaction of the opponent in different circumstances.
In the original defense location, Player B returns a net shot, which leaves Player A in a passive situation, and thus has fewer choices for returning the shuttle.
On the other hand, after changing the defense position, Player A is close to the net and makes Player B return a clear that enables Player A to launch an attack.
This verifies that the opponent returns the shuttle based on the defense location of the player; therefore, standing at an effective position blocks the choices and forces the opponent to return a predictable stroke.
With the help of DyMF providing likely strategies of returning shot types and moving towards a certain position on the court, badminton coaches and players are able to analyze opponents' implicit tactics at the stroke-level, and train for specific scenarios.


\section{Conclusion and Future Works}
\label{sec:conclusion}

In this paper, we introduce the movement forecasting task, which plays a vital role in investigating offense and defense behaviors in sports analytics.
To address the problem, we represent badminton rallies with PM graphs to delve into the strategic structure based on the interrelations between players, and propose DyMF with interaction style extractors for extracting the mutual interactions across time and capturing the dynamic tactics based on the players' past movement patterns.
Furthermore, the hierarchical fusions are designed to fuse mutual influence of players' styles and the importance of individual style and interactions.
Extensive experiments on the real-world dataset empirically show the effectiveness of our proposed model compared to the baseline methods in different scenarios.
Besides, an analysis scenario is illustrated to show the practicality of movement forecasting.
For future research, we will extend DyMF to badminton doubles for movements investigation with greater concentration on the interactions between players.
\section{Acknowledgments}

This work was supported by the Ministry of Science and Technology of Taiwan under Grants MOST-110-2627-H-A49-001.



\appendix

\begin{appendices}

\section{Related Work}
\label{sec:related-work}

\subsection{Sports Analytics}
Sports analytics aims at applying data-driven methods to a wide variety of sources of sports data for gaining insights into the tactics of teams and players.
The applications of sports analytics can be mainly divided into two categories: on-field analytics and off-field analytics.
On-field analytics focuses on improving the performance of a team or individual player by providing objective perspectives such as the fitness of players \cite{rydzik2021physical, ramirez2021effects}, player performance \cite{Wang2020badminton,jamil2021using,terner2021modeling}, and the potential rising stars \cite{DBLP:journals/kbs/MahmoodDA21}. 
On the other hand, off-field analytics assists team owners in increasing profits by studying the relationship with fans \cite{paek2021examining, yim2021identifying, qian2022watching}, and their marketing strategies \cite{melton2021examining, richelieu2021strategic}.
In this paper, we focus on on-field analytics for investigating the performance of players.
\citet{DBLP:conf/ijcai/DecroosBHD20} present five main issues around artificial intelligence in sports analytics:
representation, interpretability, decision making, understanding behavior, and experimental evaluation.
In badminton, there are only a few applications tackling these issues.
\citet{DBLP:conf/icdm/WangCYWFP21} formalized the badminton language for describing the process of a rally, which can be used for our inputs and outputs of movement forecasting (i.e., players, shot types, and locations).
ShuttleNet is proposed for stroke forecasting to simulate players' strategies based on the past information \cite{DBLP:conf/aaai/WangSCP22}.
Our work introduces another instrumental but challenging application by predicting the next movements and the shot type of both players in a rally, which helps players investigate the effect and behavior of each movement.

\subsection{Graph Neural Networks}
Graph neural networks are applied to leverage the structure of the data, e.g., molecules \cite{DBLP:conf/nips/GasteigerBG21} and social networks \cite{DBLP:conf/www/ZhangGPH22}, to learn the representation of entities based on neighborhood properties.
Typically, a graph is composed of nodes and edges, which is different from sequences like sentences in natural language.
In addition to learning node embeddings solely on the homogeneous graph structure \cite{DBLP:conf/iclr/KipfW17}, graph neural networks have been explored for various types of graphs by considering different relations between nodes \cite{DBLP:conf/esws/SchlichtkrullKB18}, temporal information \cite{DBLP:conf/cvpr/SiC0WT19, DBLP:conf/aaai/ParejaDCMSKKSL20}, and heterogeneous graphs learning (which have multiple types of nodes and edges) \cite{DBLP:conf/cikm/ZhangK0WWZY21}.
In trajectory prediction, NRI \cite{DBLP:conf/icml/KipfFWWZ18} infers the underlying relation with static assumption between entities, and predicts the future location by applying graph neural networks.
dNRI \cite{DBLP:conf/cvpr/GraberS20a} dynamically models the relations for each time step by extending the NRI framework to address changing relations between entities.
However, these previous approaches failed to model multifaceted interactions between players and players themselves at different time steps.
Our novel approach, in contrast to the existing methods, develops an interaction style extractor to capture mutual interactions through strategic relations and hierarchical fusion to fuse multifaceted interactions.
\section{Details and Setups of Baselines}
\label{implemetation-details}

\subsection{Baselines}
The details of the baseline methods are depicted as follows:
\begin{enumerate}
    \item Sequence-based models:
        \begin{itemize}
            \item Seq2Seq \cite{DBLP:conf/nips/SutskeverVL14} uses two LSTMs as encoder and decoder to enable different output lengths from the input lengths for machine translation.
            \item TF \cite{DBLP:conf/icpr/GiuliariHCG20} extends Transformers \cite{DBLP:conf/nips/VaswaniSPUJGKP17} to trajectory prediction, which models the pedestrian individually without auxiliary interaction with the environment.
            \item ShuttleNet \cite{DBLP:conf/aaai/WangSCP22} fuses the contexts of rally progress and player styles extracted by Transformer-based models to predict where to hit and which shot type it is.
        \end{itemize}
    \item Graph-based models:
        \begin{itemize}
            \item dNRI \cite{DBLP:conf/cvpr/GraberS20a} dynamically models the relation between entities that changes over time for trajectory prediction.
            \item GCN\textsubscript{PM} is a variant of GCN \cite{DBLP:conf/iclr/KipfW17}, which generates the embedding of entities in a graph by aggregating the information from neighbors.
            \item R-GCN\textsubscript{PM} is a variant of R-GCN \cite{DBLP:conf/esws/SchlichtkrullKB18}, which learns the embedding of entities in a multi-graph by aggregating the message from neighbors based on the relations.
            \item E-GCN\textsubscript{PM} is a variant of EvolveGCN \cite{DBLP:conf/aaai/ParejaDCMSKKSL20}, which extends GCN from updating the weights of GCN using recurrent neural networks to model the dynamism of the graph.
        \end{itemize}
\end{enumerate}

\subsection{Implementation Details}
In graph-based models, we regard the shot types as the edges between nodes, and the nodes represent the locations of each player for constructing a PM graph\footnote{We note that dNRI is not implemented with the PM graph since the number of entities in dNRI would not change over time.}.
We concatenate the shot type and locations of both players at each time step as the inputs for representing the current information, and take a rally as a sequence to sequence-based models.

\subsection{Hyper-parameters Settings}
Table \ref{tab:hyperparameters-settings} summarizes the details of the hyper-parameters.
The hyper-parameters were searched by conducting 5-fold cross-validation.
The batch size was chosen from \{16, 32, 48, 64\} and the dimension of embeddings ($d_l, d_p, d_e$) was chosen from \{4, 16, 32\}.
The number of layers ($l$) was chosen from \{1, 2, 3\} and the number of basis matrices ($B$) was chosen from \{1, 2, 3, 4, 5\}.
The dropout rate was chosen from \{0, 0.1, 0.2, 0.3\}.

\begin{table}
  \centering
  \begin{tabular}{ccc}
    \toprule
    Hyper-parameters                 & Value \\
    \midrule
    batch size                      & 32    \\
    hidden size                     & 16    \\
    learning rate                   & 0.001 \\
    player embedding dimension      & 16    \\
    shot type embedding dimension   & 16    \\
    location embedding dimension    & 16    \\
    epochs                          & 100   \\
    dropout rate                    & 0.1   \\
    number of basis matrix          & 3     \\
    number of sample sequence       & 10    \\
    \bottomrule
\end{tabular}
  \caption{Hyper-parameters settings.}
  \label{tab:hyperparameters-settings}
\end{table}

\subsection{Details of Evaluation Metrics} The details of the evaluation metrics following \cite{DBLP:conf/aaai/WangSCP22} are introduced as follows:

\textbf{Cross-entropy (CE)} evaluates the uncertainty of the prediction of the shot type, which is widely used to deal with the measurement of the generative model:
\begin{equation}
    CE = - \mathlarger{\mathlarger{\Sigma}}_{R \in \mathcal{D}} \mathlarger{\mathlarger{\Sigma}}^{|R|-1}_{t=\tau} s^{t} log(softmax(\hat s^{t})).
\end{equation}

\textbf{Mean square error (MSE)} calculates the square of the discrepancy of the locations between the ground truth and prediction, which is widely used in trajectory prediction: 
\begin{equation}
\begin{gathered}
    MSE = \mathlarger{\mathlarger{\Sigma}}_{R \in \mathcal{D}} \mathlarger{\mathlarger{\Sigma}}^{|R|}_{t=\tau+1} [(\hat l^{t}_{ax} - l^{t}_{ax})^2 + \\
    (\hat l^{t}_{ay} - l^{t}_{ay})^2 + (\hat l^{t}_{bx} - l^{t}_{bx})^2 + (\hat l^{t}_{by} - l^{t}_{by})^2].
\end{gathered}
\end{equation}

\begin{table}
    \small
    \centering
    \newcommand{\specialcell}[2][c]{%
  \begin{tabular}[#1]{@{}c@{}}#2\end{tabular}}

\begin{tabular}{c|ccc|c}
    \toprule
    Model                 & MSE    & MAE    & CE     & \specialcell{Avg.\\Improvement} \\
    \midrule
    w/o dynamic           & 1.1838 & 1.6389 & 1.9650 & -4.62\% \\
    w/o player-player     & 1.1346 & 1.6086 & 1.9607 & -2.40\% \\
    w/o rally             & 1.1375 & 1.6113 & 1.9883 & -3.01\% \\
    w/o style             & 1.1270 & 1.6034 & 1.9674 & -2.16\% \\
    \midrule
    DyMF (Ours)           & \textbf{1.0827} & \textbf{1.5739} & \textbf{1.9570} & - \\ 
    \bottomrule
\end{tabular}
    \caption{Ablative experiments of our model.}
    \label{tab:model-ablation}
\end{table}

\textbf{Mean absolute error (MAE) } calculates the absolute value of the discrepancy of the locations between the ground truth and prediction, which is widely used in trajectory prediction:

\begin{equation}
\begin{gathered}
    MAE = \mathlarger{\mathlarger{\Sigma}}_{R \in \mathcal{D}} \mathlarger{\mathlarger{\Sigma}}^{|R|}_{t=\tau+1} (|\hat l^{t}_{ax} - l^{t}_{ax}| + |\hat l^{t}_{ay} - l^{t}_{ay}|\\
    + |\hat l^{t}_{bx} - l^{t}_{bx}| + |\hat l^{t}_{by} - l^{t}_{by}|).
\end{gathered}
\end{equation}

\section{Additional Experimental Results}

\subsection{Ablation Study}
To analyze the contributions of different modules of DyMF, we conduct ablative experiments by removing each module with observed sequence length $\tau=8$.
In addition, we also conduct extensive experiments on different graph construction methods to verify the strength of the PM graph.

\subsubsection{Effects of Each Module.}
We testify four variants of DyMF to study the contributions of our proposed modules:
1) \textbf{w/o dynamic}: DyMF replacing dynamic GCN with GCN, 2) \textbf{w/o player-player}: DyMF removing the player-player fusion and using the outputs of dynamic GCN for player-rally fusion, 3) \textbf{w/o rally}: DyMF removing the dynamic weights of the rally interactions ($\beta_{a|b}$) in the player-rally fusion, and 4) \textbf{w/o style}: DyMF removing the dynamic weights of the player-player styles ($\alpha_{a|b}$) in the player-rally fusion.
The performance of all variant models is reported in Table \ref{tab:model-ablation}.

We observe that all modules contribute to both shot type performance and location performance, which demonstrates the effective design of each module for predicting shot types and movements of players.
Moreover, the performance of w/o dynamic drops significantly in terms of MSE for 9.6\%, which signifies that statically modeling different rally progress and different players hampers the model performance.
Our proposed dynamic GCN, in contrast, flexibly adjusts to various previous rallies and dynamically generates weights for representing the current style of each player.
Removing the player-player fusion causes the model to neglect the adaptability of players and only capture the style of an individual player, which is harmful to the performance.
The declined performance also suggests that players change their styles to match those of their opponents as the rally progresses.
From w/o rally and w/o style, the performances of shot types and locations are degraded since the information of rally interactions or players' styles are captured with fixed weights instead of the proposed pattern-related weights.
This deleterious effect suggests that players would make a decision based on the current situation and individual style with different importance at each stroke.

\begin{table}
    \small
    \centering
    \newcommand{\specialcell}[2][c]{%
  \begin{tabular}[#1]{@{}c@{}}#2\end{tabular}}
  
\begin{tabular}{c|ccc|c}
    \toprule
    Model                         & MSE    & MAE    & CE     & \specialcell{Avg.\\Improvement} \\
    \midrule
    GCN\textsubscript{C}          & 3.2579 & 2.9409 & 2.0395 & - \\
    GCN\textsubscript{PM}         & 1.2676 & 1.7508 & 1.9875 & 75.87\% \\
    \midrule
    R-GCN\textsubscript{C}        & 1.1296 & 1.6211 & 1.9758 & - \\ 
    R-GCN\textsubscript{PM}       & 1.2006 & 1.6554 & 2.0470 & -3.82\% \\
    \midrule
    DyMF\textsubscript{C} (Ours)  & 1.1366 & 1.6277 & 1.9811 & - \\
    DyMF (Ours)                   & \textbf{1.0827} & \textbf{1.5739} & \textbf{1.9570} & 3.21\%  \\
    \bottomrule
\end{tabular}
    \caption{Performance of using PM graphs and complete graphs on graph-based models.}
    \label{tab:graph-ablation}
\end{table}

\subsubsection{Effects of Graph Construction Methods.}

To validate the effectiveness of the PM graph, we compare PM graphs with complete graphs on graph-based methods.
We use a complete graph directly for GCN (GCN\textsubscript{C}), and a dummy relation is added to a PM graph to form a complete graph for R-GCN (R-GCN\textsubscript{C}) and our model (DyMF\textsubscript{C}).
We remove E-GCN\textsubscript{PM} as the unstable performance, which would bias the performance when applying complete graphs.
In Table \ref{tab:graph-ablation}, we observe that GCN\textsubscript{C} has a significant drop compared to using PM graphs (GCN\textsubscript{PM}).
These results show that a complete graph connects irrelevant information for each time step, which causes noises for learning relations (i.e., players rarely consider the first stroke when the rally lasts for multiple strokes.).
On the other hand, our proposed PM graph propagates information to the adjacent time steps, which is more important to make decisions at the next time step.
We note that R-GCN\textsubscript{C} performs slightly better than using complete graphs, which is because using complete graphs with relations is similar to our dynamic GCN, which enables R-GCN\textsubscript{C} to differentiate information between nodes.
Meanwhile, our model still outperforms these baselines equipped with PM graphs, which showcases the strength of our model's capacity.




\subsection{Extra Case Study: Different Strategies Against the Same Player}

\begin{figure}
    \centering
    \includegraphics[width=.91\linewidth]{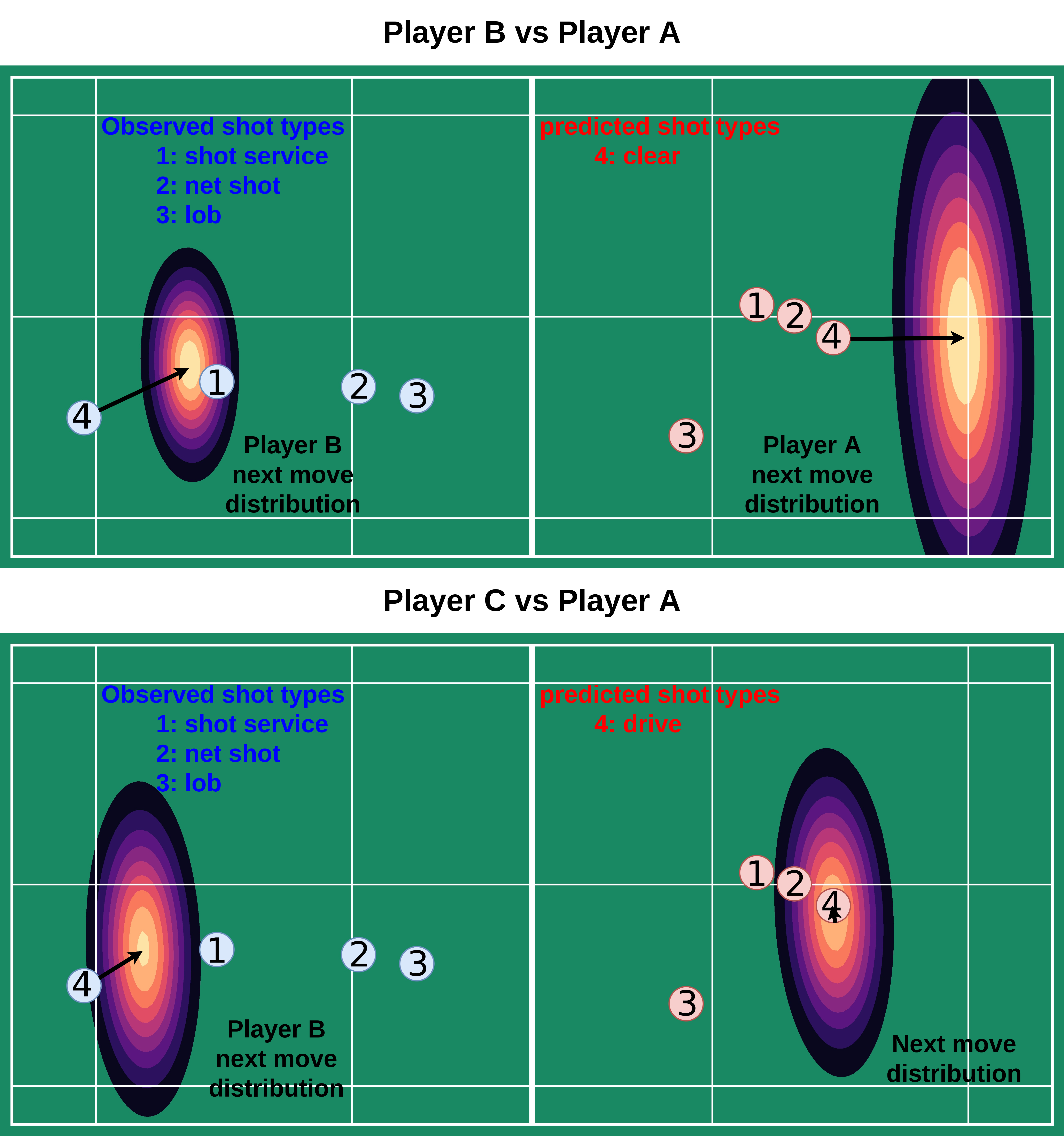}
    \caption{Illustrations of different strategies against the same player. Red nodes are the serving player (Player A) and blue nodes are the receiving players (Player B and Player C).}
    \label{fig:case-study-different-player}
\end{figure}

In addition to strategies in different defense locations, we also conducted an analysis scenario of different strategies against the same player.
Figure \ref{fig:case-study-different-player} depicts the scenario of different matchups: Player A vs Player B (a) and Player A vs Player C (b).
After three strokes, Player B/C (blue node) is going to return the stroke and defend; meanwhile, Player A (red node) is going to move and return the shuttle.
We can observe that Player B is likely to return a clear to mobilize Player A to the back court.
On the other hand, Player C is more likely to return a drive straight to Player A, which is more aggressive than Player B.
It is worth noting that a clear has a longer flying time, which gives Player B more time to move to the defense position, while returning a drive has a shorter moving distance.
This case demonstrates that our DyMF is able to reveal not only the strategies of defending locations but also the strengths and weaknesses of players, which provides multifaceted usage of movement forecasting.

\end{appendices}



\bibliography{reference}

\end{document}